\documentclass[sigconf]{acmart}
\AtBeginDocument{%
  \providecommand\BibTeX{{%
    Bib\TeX}}}

\usepackage{multirow,enumitem} 
\usepackage{mathtools} 
\usepackage{booktabs} 
\usepackage{hyperref} 
\usepackage{pgfplots}
\usepackage{subcaption}
\usepackage{algorithmic} 

\usepackage[linesnumbered,ruled,vlined]{algorithm2e}

\SetCommentSty{mycommfont}

\usetikzlibrary{pgfplots.groupplots}
\usetikzlibrary{patterns}
\setlist[itemize]{noitemsep, topsep=0pt}

\newenvironment{customlegend}[1][]{%
  \begingroup
  \csname pgfplots@init@cleared@structures\endcsname
  \pgfplotsset{#1}%
}{%
  \csname pgfplots@createlegend\endcsname
  \endgroup
}%
\def\addlegendimage{\csname pgfplots@addlegendimage\endcsname}

\def\BibTeX{{\rm B\kern-.05em{\sc i\kern-.025em b}\kern-.08em
    T\kern-.1667em\lower.7ex\hbox{E}\kern-.125emX}}

\copyrightyear{2026}
\acmYear{2026}
\setcopyright{cc}
\setcctype{by}
\acmConference[KDD '26]{Proceedings of the 32nd ACM SIGKDD Conference on Knowledge Discovery and Data Mining V.1}{August 09--13, 2026}{Jeju Island, Republic of Korea}
\acmBooktitle{Proceedings of the 32nd ACM SIGKDD Conference on Knowledge Discovery and Data Mining V.1 (KDD '26), August 09--13, 2026, Jeju Island, Republic of Korea}
\acmPrice{}
\acmDOI{10.1145/3770854.3780309}
\acmISBN{979-8-4007-2258-5/2026/08}

\newcommand{\agnes}{\textsf{AGNES}}

\usepackage{pdfrender}

\newcounter{mysfig}
\counterwithin{mysfig}{figure}

\renewcommand\themysfig{(\alph{mysfig})}
\makeatletter
\newcommand\Scaption[1]{%
\refstepcounter{mysfig}%
\vskip.5\abovecaptionskip
  \sbox\@tempboxa{\small\themysfig~#1}%
  \ifdim \wd\@tempboxa >\hsize
    \small\themysfig~#1\par
  \else
    \global \@minipagefalse
    \hb@xt@\hsize{\hfil\box\@tempboxa\hfil}%
  \fi
  \vskip\belowcaptionskip}
\makeatother

\newcommand{\codeurl}{\url{https://github.com/Bigdasgit/agnes-kdd26}}
\begin{document}

\title{Accelerating Storage-based Training for Graph Neural Networks}


\author{Myung-Hwan Jang}
\affiliation{%
  \institution{Hanyang University}
  \city{Seoul}
  \country{Republic of Korea}
}
\email{sugichiin@hanyang.ac.kr}
\orcid{0000-0003-4419-5148}

\author{Jeong-Min Park}
\affiliation{%
  \institution{Hanyang University}
  \city{Seoul}
  \country{Republic of Korea}}
\email{jmpark96@hanyang.ac.kr}
\orcid{0000-0001-9389-6501}

\author{Yunyong Ko}
\affiliation{%
  \institution{Chung-Ang University}
  \city{Seoul}
  \country{Republic of Korea}
}
\email{yyko@cau.ac.kr}
\orcid{0000-0003-1283-4697}

\author{Sang-Wook Kim}
\authornote{Corresponding author.}
\affiliation{%
  \institution{Hanyang University}
  \city{Seoul}
  \country{Republic of Korea}}
\email{wook@hanyang.ac.kr}
\orcid{0000-0002-6345-9084}

\renewcommand{\shortauthors}{Myung-Hwan Jang, Jeong-Min Park, Yunyong Ko, and Sang-Wook Kim}

\begin{abstract}
Graph neural networks (GNNs) have achieved breakthroughs in various real-world downstream tasks due to their powerful expressiveness. 
As the scale of real-world graphs has been continuously growing, 
\textit{a storage-based approach to GNN training} has been studied, which leverages external storage (e.g., NVMe SSDs) to handle such web-scale graphs on a single machine.
Although such storage-based GNN training methods have shown promising potential in large-scale GNN training, 
we observed that they suffer from a severe bottleneck in data preparation since they overlook a critical challenge:
\textit{how to handle a large number of small storage I/Os}.
To address the challenge, in this paper, we propose a novel storage-based GNN training framework, named \textsf{AGNES},
that employs a method of \textit{block-wise storage I/O processing} to fully utilize the I/O bandwidth of high-performance storage devices.
Moreover, to further enhance the efficiency of each storage I/O, 
\textsf{AGNES} employs a simple yet effective strategy, \textit{hyperbatch-based processing} based on the characteristics of real-world graphs.
Comprehensive experiments on five real-world graphs reveal that
\textsf{AGNES} consistently outperforms four state-of-the-art methods, up to 4.1$\times$ faster than the best competitor.
\end{abstract}



\maketitle
\newcommand\kddavailabilityurl{https://doi.org/10.5281/zenodo.18112452}
\ifdefempty{\kddavailabilityurl}{}{
\begingroup\small\noindent\raggedright\textbf{Resource Availability:}\\
The source code of this paper has been made publicly available at \url{\kddavailabilityurl} ({\codeurl}).
\endgroup
}

\section{Introduction}
\label{sec:introduction}

\textit{Graphs} are prevalent in many applications~\cite{Kyr12, Roy13, Han13, Zhe15,Jo19,Jan23} to represent a variety of real-world networks,
such as social networks and web,
where objects and their relationships are modeled as nodes and edges, respectively.
Recently, \textit{graph neural networks} (GNNs), a class of deep neural networks specially designed to learn such graph-structured data, 
have achieved breakthroughs in various downstream tasks, including node classification ~\cite{Ron19, Zho19}, link prediction ~\cite{Zha18,Yu21,Yin18,Pal20,Ya21}, and community detection ~\cite{Luo21, Gao21, Zha23b}.

\begin{figure}[t]
		\centering
		\includegraphics[width=0.47\textwidth]{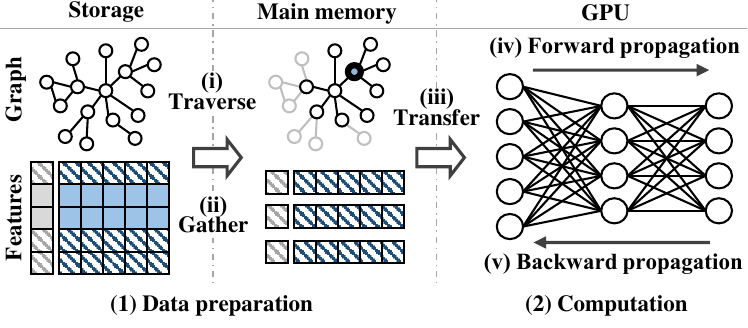}
		\vspace{-4mm}
		\caption{Overview of storage-based GNN training.}
		\vspace{-6mm}
	\label{fig:gnn_overview}
\end{figure}

Although existing works have designed model architectures to learn the structural information of graphs by considering not only \textit{node features} but also \textit{graph topology}~\cite{Liu21, Rey23, Li19, Erg23, Zha23a},
they have a \textit{simple assumption}: 
the entire input data, including node features and graph topology, \textit{reside in the GPU or main memory} during GNN training~\cite{Fey19, Wan19, Wu20, Zha20}.
However, as the scale of real-world graphs has been continuously growing,
this assumption is not practical anymore:
the size of real-world graphs often exceeds the capacity of GPU memory (e.g., 80GB for an NVIDIA H100) or even that of main memory in a single machine (e.g., 256GB).
For instance, training 3-layer GAT~\cite{Vel18} on the yahoo-web graph~\cite{Yah24}, which consists of 1.4B nodes and 6.6B edges, requires about 1.5TB, including node features, graph topology, and intermediate results.

\begin{figure*}[t]
	\begin{tikzpicture}
    \begin{customlegend}[legend columns=2, legend style={align=center, draw=none, at={(4.6,2.6)}, column sep=0.1ex, row sep=-0.5ex, font=\footnotesize}, legend entries={SAGE,GCN}]
    \addlegendimage{area legend, red, fill=none, postaction={pattern = north west lines, pattern color = red}, line width=0.5pt}
    \addlegendimage{area legend, blue, fill=none, postaction={pattern = north east lines, pattern color = blue}, line width=0.5pt}
    \end{customlegend}
	\begin{axis}[
	footnotesize,
	height=3.5cm,
	width=0.26\textwidth, 
	axis x line*=center,
	axis y line*=left,
	ybar, 
    bar width=8pt, 
	ymajorgrids=true,
	enlarge x limits=0.35,
	title={(a) Execution time breakdown},
	title style={xshift=50pt,yshift=-88pt},
	xlabel style={yshift=4pt},
	ylabel={Ratio of data prep. (\%)},
	ylabel style={yshift=-13pt},
    xlabel=Ginex,
    ymin=0, ymax=100,
	ytick={0,20,40,60,80,100},
	symbolic x coords={start,TW,PA,FR,end}, 
	xtick=data,
	legend style={at={(0.4,1.1)}, anchor=south, draw=black, legend columns=2},
	every node near coord/.append style={color=black, font=\small}
	]
    \addplot+[ybar, pattern = north west lines, area legend, pattern color = red, draw = red]
	coordinates {
		(TW, 90) 
		(PA, 95) 
		(FR, 97) 
	};

    \addplot+[ybar, pattern = north east lines, area legend, pattern color = blue, draw = blue] 
	coordinates {
		(TW, 86) 
		(PA, 92) 
		(FR, 96) 
	};
	\end{axis}
    \hspace{37mm}
    \begin{axis}[
	footnotesize,
	height=3.5cm,
	width=0.26\textwidth, 
	axis x line*=center,
	axis y line*=left,
	ybar, 
    bar width=8pt, 
	ymajorgrids=true,
	enlarge x limits=0.35,
    title style={yshift=-80pt},
	xlabel style={yshift=4pt},
	ylabel style={yshift=-13pt},
    xlabel=GNNDrive,
    ymin=0, ymax=100,
	ytick={0,20,40,60,80,100},
	symbolic x coords={start,TW,PA,FR,end}, 
	xtick=data,
	legend style={at={(0.4,1.1)}, anchor=south, draw=black, legend columns=2},
	every node near coord/.append style={color=black, font=\small}
	]
    \addplot+[ybar, pattern = north west lines, area legend, pattern color = red, draw = red]
	coordinates {
		(TW, 90) 
		(PA, 88) 
		(FR, 85) 
	};

    \addplot+[ybar, pattern = north east lines, area legend, pattern color = blue, draw = blue] 
	coordinates {
		(TW, 87) 
		(PA, 86) 
		(FR, 85) 
	};
	\end{axis}
    
	\end{tikzpicture}
    \hspace{22mm}
	\begin{tikzpicture}
	\begin{axis}[
	footnotesize,
	height=3.5cm,
	width=0.26\textwidth, 
	axis x line*=bottom,
	axis y line*=none,
	ybar,
    bar width=7pt, 
	ymajorgrids=true,
	enlarge x limits=0.15, 
    major grid style={line width=.2pt,draw=gray!50},
	title={(b) Dist. of storage I/O sizes},
	title style={yshift=-88pt},
	xlabel style={yshift=4pt},
    ylabel =Ratio of storage I/Os (\%),
    ylabel style={yshift=-12pt, align=center, text width=3cm},
    xlabel=I/O size (KB),
    ymin=0, ymax=100,
	xtick=data,
	symbolic x coords={4, 8, 16, 32, 64, 128, 256, 512},
    ytick={0, 20, 40, 60, 80, 100},
    xticklabel style=,
    nodes near coords,
    every node near coord/.append style={
        color=black, font=\footnotesize,
        rotate=90,
        anchor=center,
        xshift=9pt,
        yshift=0pt,
        /pgf/number format/fixed,
        /pgf/number format/precision=3,
    },
	]	
\addplot+[ybar, pattern = north east lines, area legend, pattern color = red, draw = red, point meta=y,	visualization depends on=rawy\as\rawy, nodes near coords={\pgfmathprintnumber{\rawy}}] 
	coordinates {
        (4,75)
        (8,19)
        (16,4.7)
        (32,0.62)
        (64,0.19)
        (128, 0.031)
        (256, 0.0019)
	};
	\end{axis}
    \end{tikzpicture}
    \centering
    \begin{tikzpicture}
    \begin{customlegend}[legend columns=1, legend style={align=center, draw=none, at={(2.7,1.9)}, column sep=0.1ex, row sep=-0.5ex, font=\footnotesize}, legend entries={Ginex,GNNDrive}]
    \addlegendimage{area legend, red, fill=none, postaction={pattern = north west lines, pattern color = red}, line width=0.5pt}
    \addlegendimage{area legend, blue, fill=none, postaction={pattern = north east lines, pattern color = blue}, line width=0.5pt}
    \end{customlegend}
	\begin{axis}[
	footnotesize,
	height=3.5cm,
	width=0.26\textwidth, 
	axis x line*=center,
	axis y line*=left,
	ybar, 
    bar width=8pt, 
	ymajorgrids=true,
	enlarge x limits=0.35,
	title={(c) Resource utilization},
	title style={yshift=-88pt},
	xlabel style={yshift=4pt},
	ylabel={GPU util. (\%)},
	ylabel style={yshift=-18pt},
    xlabel=Datasets,
    ymin=0, ymax=100,
	ytick={0,20,40,60,80,100},
	symbolic x coords={start,TW,PA,FR,end}, 
	xtick=data,
	legend style={at={(0.4,1.1)}, anchor=south, draw=black, legend columns=2},
	every node near coord/.append style={color=black, font=\small}
	]
    \addplot+[ybar, pattern = north west lines, area legend, pattern color = red, draw = red]
	coordinates {
		(TW, 10) 
		(PA, 25) 
		(FR, 6) 
	};

    \addplot+[ybar, pattern = north east lines, area legend, pattern color = blue, draw = blue] 
	coordinates {
		(TW, 9) 
		(PA, 22) 
		(FR, 4) 
	};
	\end{axis}	
	\end{tikzpicture}
    
    \vspace{-4mm}
	\caption{Breakdown of the execution time of state-of-the-art GNN training methods (Ginex~\cite{Par22} and GNNDrive~\cite{Jia24}).}
\label{fig:introginex_horizon}
\vspace{-5mm}
\end{figure*}
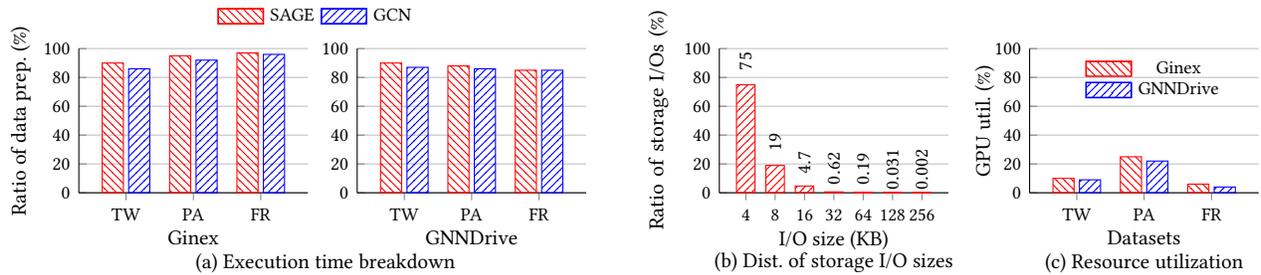

To address this challenge, 
a \textit{storage-based approach to GNN training} has been studied~\cite{Par22,Wal23,Jia24,She24},
which leverages recent high-performance \textit{external storage devices} (e.g., NVMe SSDs)~\cite{Par22,Wal23,Jia24}.
This approach stores the entire graph topology and node features in external storage and loads some \textit{parts} into the main memory from storage only when required for GNN training.
The storage-based GNN training is two-fold as illustrated in Figure~\ref{fig:gnn_overview}:

\vspace{0.5mm}
\noindent
(1) \textbf{Data preparation}: it (i) traverses the graph stored in storage (by loading it into main memory) to find the neighboring nodes of \textit{target nodes} necessary for training,
(ii) gathers their associated features stored in storage for training in main memory, and (iii) transfers both into the GPU.

\vspace{0.5mm}
\noindent
(2) \textbf{Computation}: it performs (iv) forward propagation (i.e., prediction) and (v) backward propagation (i.e., loss and gradient computations) over the transferred data in the GPU.

Although the advanced computational power of modern GPUs has accelerated the computation stage significantly, 
the data preparation stage could be a significant bottleneck in the entire process of the storage-based GNN training as it can incur a large amount of \textit{I/Os between storage and main memory} (simply storage I/Os, hereafter).
Existing works~\cite{Par22,Wal23,Jia24,She24} have focused on improving the data preparation stage, thus showing its promising potential.

Despite their success, we observed that there is still a large room for further improvement in storage-based GNN training.
We conducted a preliminary experiment to analyze the ratio of the time for the data preparation stage to the total execution time in Ginex~\cite{Par22} and GNNDrive~\cite{Jia24},  state-of-the-art methods for storage-based GNN training.
Specifically, we trained two GNN models, i.e., GCN~\cite{Kip16} and GraphSAGE~\cite{Ham17} (SAGE in short), on three real-world graph datasets -- twitter-2010 (TW)~\cite{Jur14}, ogbn-papers100M (PA)~\cite{Hu20}, and com-friendster (FR)~\cite{Jur14}.
As shown in Figure~\ref{fig:introginex_horizon}(a), 
\textit{the data preparation stage dominates the entire training process} (i.e., up to 96\% of the total execution time).
For in-depth analysis, we also measured the size of each individual I/O that occurs during training.
Figure~\ref{fig:introginex_horizon}(b) shows the distribution of storage I/Os' sizes, 
where \textit{a large number of storage I/Os are small}, while only a few I/Os are very large.
Such a large number of small I/Os leads to significant degradation of the utilization of computing resources (e.g., GPU utilization) in the GNN training as shown in Figure~\ref{fig:introginex_horizon}(c).

We posit that this phenomenon arises because real-world graphs tend to have a \textit{power-law degree distribution}~\cite{Les07}, meaning that the majority of nodes have only a few edges (i.e., neighbors) while a small number of nodes have a huge number of edges. 
That is, the number of neighboring nodes required for GNN training is highly likely to be very small in most cases.
Existing storage-based GNN training methods~\cite{Par22,Wal23,Jia24,She24}, however, have overlooked this important characteristic.
They focus only on \textit{how to increase the possibility of reusing cached data in main memory} (i.e., cache hit ratio) and simply read a few nodes from storage whenever they are required for GNN training, 
thereby \textit{generating a significant number of small storage I/Os}.
For example, 
\citet{She24} aims to enhance the locality of sampled nodes for better cache hit ratio by partitioning the entire graph and selecting target nodes within the same partition. 
These approaches, however, do not address the challenge of handling a large number of small I/Os fundamentally,
which still remains under-explored.

We may tackle this challenge by merging small storage I/Os and processing them together.
However, simply increasing the I/O unit size is not an optimal way to solve the problem since a large amount of unnecessary data (i.e., nodes/edge unrelated to target nodes) can be included in each I/O, which can waste the main memory space.

\vspace{0.5mm}
\noindent
\textbf{Our work}. 
In this paper, to address the aforementioned fundamental challenge of storage-based GNN training, 
we propose a novel framework, named as \textbf{{\agnes}}, that has a 3-layer architecture with (i) storage, (ii) in-memory, and (iii) operation layers where each layer interacts closely with the others for efficiently handling storage I/Os.
We present a method of \textit{block-wise storage I/O processing} with a \textit{novel data layout} to reduce the number of small storage I/Os, thereby fully utilizing the power of high-performance storage devices--- the I/O bandwidth (I/O-BW).
Moreover, to further improve the efficiency of each storage I/O (i.e., the cache hit ratio), 
we propose a simple yet effective strategy based on the characteristics of real-world graphs: 
\textit{hyperbatch-based processing}, which carefully collects the data required for GNN training within blocks as much as possible and process them all at once in each iteration.


\vspace{1mm}
\noindent
\textbf{Contributions}. The main contributions of this work are as follows.
\begin{itemize}[leftmargin=10pt]
    \item \textbf{Observations}: We observe that existing works have overlooked a critical yet under-explored challenge of storage-based GNN training: \textit{how to handle a large number of small storage I/Os}.
        
    \item \textbf{Framework}: We propose a novel framework for storage-based GNN training, {\agnes}, that effectively addresses the challenge by employing block-wise storage I/O processing and hyperbatch-based processing.
    
    \item \textbf{Evaluation}: 
    Comprehensive experiments using five real-world graphs reveal that {\agnes} significantly outperforms state-of-the-art storage-based GNN training methods. Specifically, {\agnes} finishes training by up to 4.1$\times$ faster, while achieving the utilization of I/O-BW by up to 4.5$\times$ greater than the best competitor.
\end{itemize}

\vspace{-5mm}
\section{Related Works}\label{sec:related}
In this section, we review existing GNN training approaches and explain their relation to our work.

\vspace{1mm}
\noindent
\textbf{Storage-based approaches}.
Recently, storage-based GNN training approaches, our main focus, have been studied, which leverage external storage on a single machine for large-scale GNN training~\cite{Par22,Wal23,Jia24,She24}.
These approaches store the entire graph topology and node features in the external storage and load only the parts required for GNN training into main memory.
Ginex~\cite{Par22}, the state-of-the-art storage-based GNN training method, 
employs a caching mechanism for node feature vectors, which addresses I/O congestion issues effectively,
thereby successfully handling billion-scale graph datasets on a single machine.
MariusGNN~\cite{Wal23} partitions the graph into multiple partitions, buffering the partitions in main memory and reusing sampled results to mitigate storage I/O bottlenecks.
It also adopts a data structure to minimize the redundancy of multi-hop sampling and the two-level minibatch replacement policy for disk-based training.
GNNDrive~\cite{Jia24} employs buffer management across different stages that support the sample stage to relieve memory contention.
It also uses an asynchronous feature extraction to address memory usage issues and alleviate storage I/O bottlenecks.
OUTRE~\cite{She24} employs partition-based batch construction and historical embedding to reduce neighborhood redundancy and temporal redundancy in sampling-based GNN training.

Although the existing storage-based approaches have shown promising potential for large-scale GNN training on a single machine,
they still suffer from a significant bottleneck in the data preparation stage (as shown in Figure~\ref{fig:introginex_horizon}) since they overlook the natural but critical challenge.
To the best of our knowledge, 
this is the first work to address the challenge in detail.

\vspace{1mm}
\noindent
\textbf{Other approaches}.
In addition to the storage-based approach,
memory-based and distributed-system-based approaches have been studied.
Memory-based approaches~\cite{Fey19,Wan19,Lin20,Par23,Sun23b} store graph data or node features in the main memory to handle large-scale graphs that exceed the GPU memory size.
For instance,
PyG~\cite{Fey19} employs a method of utilizing both CPU and GPU to improve the training speed of GNN models.
DGL~\cite{Wan19} adopts a zero-copy approach for fast data transfer from main memory to GPU. 
PaGraph~\cite{Lin20} addresses data transfer bottleneck between CPU and GPU by caching frequently accessed high out-degree nodes in GPU memory.
On the other hand, 
distributed-system-based approaches~\cite{Zhu19,Zhe20,Zhe22} leverage abundant computing power and memory capacity of distributed systems for training GNN models on very large graphs that even exceed the capacity of a single machine.
AliGraph~\cite{Zhu19} adopts a method of caching node data locally on each machine to reduce network communication costs.
DistDGL~\cite{Zhe20} splits a given graph using a min-cut partitioning algorithm to not only reduce network communication costs but also balance the graph partitions and minibatches generated from each partition.
DistDGLv2~\cite{Zhe22} improves DistDGL by using a multi-level partitioning algorithm and an asynchronous minibatch generation pipeline.

However, memory-based approaches cannot handle large-scale graphs that exceed the capacity of the main memory (i.e., less scalable), and distributed-system-based approaches require a substantial amount of \textit{inter-machine communication} overhead to aggregate the results from multiple machines and \textit{costs and efforts} to maintain high-performance distributed systems (i.e., less efficient and costly).

\vspace{-1.5mm}
\section{Proposed Framework: {\agnes}}\label{sec:proposed}
\vspace{-0.5mm}
In this section, we propose a novel framework for storage-based GNN training, named
\textbf{\underline{A}}ccelerating storage-based training for \textbf{\underline{G}}raph \textbf{\underline{NE}}ural network\textbf{\underline{S}} (\textbf{{\agnes}}).

\vspace{-2mm}
\subsection{Preliminaries}\label{subsec:proposed-overview}
\noindent
The notations used in this paper are described in Table~\ref{table:notations}.

\vspace{1mm}
\noindent
\textbf{Graph neural networks (GNNs)}.
GNNs aim to represent the embedding vectors of nodes based on a graph structure via a message passing mechanism~\cite{Ham17,Kip16}.
More specifically, each layer of a GNN consists of two steps: aggregation and update.
In the aggregation step, for each node, the embedding vectors of its in-neighbors are aggregated into the embedding of the target node.
In the update step, the aggregated embeddings are passed through a fully connected layer with a nonlinear function.
The update of an embedding $\textbf{h}_{v}\in\mathbb{R}^{d}$ can be represented as (1):
\begin{equation}
    \textbf{h}^{(i+1)}_{v}=\psi(\phi(\textbf{h}^{(i)}_{v'} | v' \in N(v), \textbf{h}^{i}_{v}))
\end{equation}
where $N(v)$ denotes the set of the neighboring nodes of node $v$, $\psi(\cdot)$ and $\phi(\cdot)$ are aggregation and update functions, respectively.
By stacking multiple layers, a GNN model can reflect the information of $k$-hop neighboring nodes of a target node into its embeddings, where each layer is responsible for aggregating and updating the information of neighboring nodes from the corresponding hop.

\vspace{1mm}
\noindent
\textbf{Minibatch training for GNNs}.
Meanwhile, storing a real-world graph with its node features often requires more than hundreds of gigabytes (GB) or even tens of terabytes (TB), exceeding the capacity of main memory.
To handle such a large graph on a single machine, 
a storage-based GNN training method stores the entire graph in external storage (e.g., NVMe SSDs) and processes only a subset of nodes (i.e., a \textit{minibatch}) from the entire graph, which can be loaded into GPU memory, at each iteration~\cite{Par22,Wal23}.

Even in minibatch training of GNNs, however, 
collecting all $k$-hop neighboring nodes of a target node and their feature vectors
may require a large amount of memory~\cite{Par22,Wal23}.
To address this memory issue, existing storage-based methods employ a simple strategy that (i) randomly samples only a subset of neighboring nodes related to a target node and (ii) uses them to update its embedding in GNN training~\cite{Par22,Wal23,Jia24}.

\begin{table}[t]
\small
\caption{Notations and their descriptions}
\vspace{-4mm}
\setlength\tabcolsep{4pt}
\def\arraystretch{0.86} 
\centering
\begin{tabular}{cl}
\toprule
 \textbf{Notation} & \textbf{Description}\\
\midrule
$\textbf{h}^{i}_{v}$ & an embedding of a node $v$ from $i$-th layer \\
$\psi(\cdot)$, $\phi(\cdot)$ & aggregation and update functions \\
$N(v)$ & a set of the neighboring nodes of a node $v$ \\
\midrule
$B_{g}, B_{f}$ & topology and features of input graph \\
$T^{g}_{buf}, T^{f}_{buf}$ & buffer index tables for topology $B_{g}$ and features $B_{f}$\\
$T^{g}_{obj}$ & object index table for topology $B_{g}$\\
$C_f$ & feature cache for features $B_{f}$\\
$T^f_{ch}$ & cache index table for features in feature cache $C_f$ \\
\bottomrule
\end{tabular}
\label{table:notations}
\vspace{-5mm}
\end{table}

\vspace{1mm}
\noindent
\textbf{Two stages of storage-based GNN training}.
Figure~\ref{fig:gnn_overview} shows an overview of the two main stages in storage-based GNN training: (1) data preparation and (2) computation.
In storage-based GNN training,
CPU and GPU collaborate interactively to efficiently perform large-scale GNN training~\cite{Fey19,Par22,Wal23,Jia24},
where the CPU is in charge of (1) the data preparation and the GPU is in charge of (2) the computation.
In the data preparation stage, the CPU (i) traverses a graph to find the neighboring nodes required for training, (ii) gathers their feature vectors, and (iii) transfers them to the GPU. 
In the computation stage, the GPU performs (iv) forward and (v) backward propagation (i.e., gradient computations) for each minibatch.

\begin{figure*}[t]
		\centering
		\includegraphics[width=0.95\textwidth]{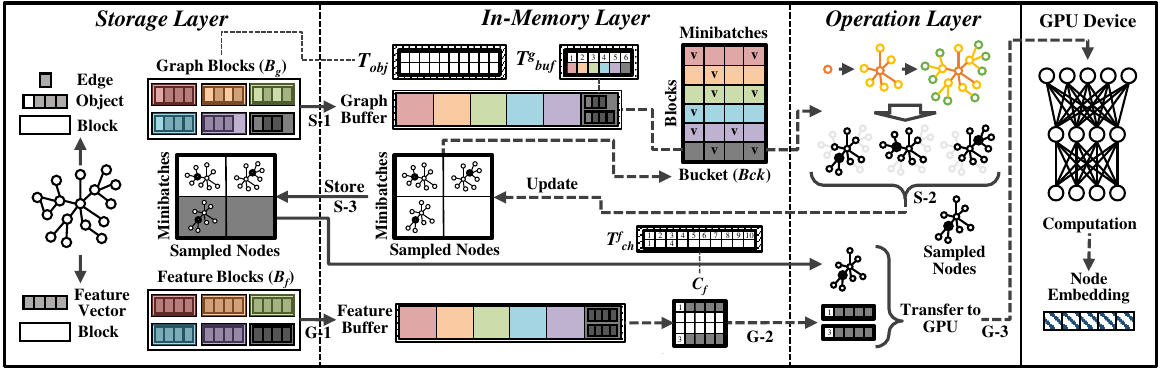}
		\vspace{-3mm}
		\caption{Overview of {\agnes} with the 3-layer architecture (Storage, In-memory, and Operation layers).}
		\vspace{-3mm}
	\label{fig:archi}
\end{figure*}

\vspace{-1.5mm}
\subsection{Architecture of {\agnes}}\label{subsec:proposed-architecture}
\vspace{-0.5mm}
{\agnes} has a 3-layer architecture with storage, in-memory, and operation layers where each layer interacts closely with the others for efficient management. 
Figure~\ref{fig:archi} shows the overview of {\agnes}.

\vspace{1mm}
\noindent
\textbf{(1) Storage layer.}
This layer manages storage space and manages graph topology and node features stored in the storage.
Specifically, it divides and stores the graph topology and feature vectors into multiple \textit{blocks} (fixed-size storage I/O unit). 
There are two types of blocks: (1) graph block and (2) feature block.
(1) A graph block contains multiple objects (i.e., multiple nodes and their related edges).
(2) A feature block contains multiple feature vectors.
If an object exceeds the size of a single block (i.e., a node with a large number of edges), the object is split across multiple blocks.
This layer is in charge of handling I/O requests from the in-memory layer, 
where all I/Os are processed in a block-wise manner.

To enhance the efficiency of block-wise storage I/O processing,
we employ an efficient data layout by following~\cite{Jo19,Jo21}.
The key idea is to place the data accessed together during an graph algorithm in the same (adjacent) blocks.
Since {\agnes} stores objects (each having a node and its edges) in blocks in the ascending order of node IDs, we assign consecutive node IDs to the nodes likely to be accessed together at the same or adjacent iteration(s) by a graph algorithm.
As a result, not only the number of accessed blocks is reduced but also the degree of sequential accesses of blocks increases, thereby enhancing the efficiency of block-wise storage I/O processing.

\vspace{1mm}
\noindent
\textbf{(2) In-memory layer.}
This layer manages the buffers in the main memory. 
Specifically, this layer is in charge of the following three tasks: (1) loading the required blocks into main memory; (2) storing the sampled results from the operation layer in the storage; and (3) gathering minibatch workloads required for GNN computation.
This layer defines the following components:
\begin{itemize}[leftmargin=10pt]
    \item Graph buffer and feature buffer: storing the graph blocks and the feature blocks loaded from storage, respectively.
    \item Buffer index tables ($T^g_{buf}$ and $T^f_{buf}$): indicating the address where the blocks are located in the graph buffer or feature buffer.
    \item Object index table ($T^g_{obj}$): mapping each block to the objects in storage, where each column corresponds to the objects stored in the corresponding block, indexed by their node IDs.
    \item Feature cache ($C_f$): storing the node features required for processing each minibatch.
    \item Cache index table ($T^f_{ch}$): tracking the location of each node’s feature in the feature cache.
\end{itemize}
To efficiently use main memory, we only store the first and last object indices for each block in the object index table, 
sorted in ascending order by node IDs.
The object index table is always pinned in the main memory to quickly identify the location of required blocks in storage.
Since this table occupies less than 0.01\% of the size of the original graph, it does not affect the overall performance.

\vspace{1mm}
\noindent
\textbf{(3) Operation layer.}
This layer is responsible for performing CPU computations involved in data preparation using the data in $B_g$ and $B_f$. 
The sampling process consists of the following steps: 
(S-1) reading the required objects from $B_g$ by referring to $T^g_{buf}$, 
(S-2) performing traversal and sampling neighboring nodes, 
and (S-3) updating the sampled nodes (target nodes and their $k$-hop neighbors) and storing them in storage. 
Then, the gathering and transferring process consists of the following steps: 
(G-1) reading the associated feature vectors of the sampled $k$-hop neighboring nodes from $B_f$ by referring to $T^f_{ch}$,
(G-2) collecting the associated feature vectors into a contiguous memory space, 
and (G-3) transferring the sampled nodes and their feature vectors to the GPU, which performs GNN computations.
This process is repeated until all minibatches are processed (e.g., one epoch).

\begin{figure}[t]
	\centering
    \vspace{-3mm}  
	\begin{tikzpicture}
    \begin{customlegend}[legend columns=1, legend style={align=center, draw=, at={(4.8,1.85)}, column sep=0.4ex, row sep=-0.5ex, font=\footnotesize}, legend entries={Cache hit ratio, Amount of I/Os}]
    \addlegendimage{red, mark=oplus, mark options ={fill=red, draw=red}, line width=0.5pt}
    \addlegendimage{area legend, gray, fill=none, postaction={pattern = north east lines, pattern color = gray}, line width=0.3pt}
    \end{customlegend}
	\begin{axis}[
	footnotesize,
	height=3.5cm,
	width=0.43\textwidth, 
	axis x line*=center,
	axis y line*=right,
	ybar,
    bar width=9pt, 
	ymajorgrids=true,
	enlarge x limits=0.1, 
    major grid style={line width=.2pt,draw=gray!50},
	xlabel style={yshift=3pt},
    ylabel style={anchor=center,rotate=180,yshift=230pt, align=center, text width=3cm},
    ylabel=Storage I/Os (TB),
    ymin=0, ymax=20000,
    ytick={0,5000, 10000, 15000, 20000},
    yticklabels={0,5, 10, 15, 20},
    scaled y ticks = false,
    yticklabel style={font=\small,},
	symbolic x coords={0, 4, 8, 16, 32, 64, 128, 256, 512, 1024},
    xlabel=Storage I/O unit size (KB),
    xtick=data,
    xticklabel style={anchor=near xticklabel, yshift=0pt}, 
    legend style={at={(0.5,1.6)}, anchor=north, draw=none, legend columns=-1},
	every node near coord/.append style={color=black, font=\small}
    ]
    every node near coord/.append style={
        color=black, font=\small
    }
	]	
\addplot+[ybar, pattern = north east lines, area legend, pattern color = gray, draw=gray, point meta=y, visualization depends on=rawy\as\rawy] 
	coordinates {
		(4, 61)
		(8, 122)
		(16, 245)
		(32, 490)
		(64, 981)
		(128, 1963)
		(256, 3928)
		(512, 7855)
            (1024, 15710)
	};

	\end{axis}

	\begin{axis}[
	footnotesize,
	height=3.5cm,
	enlarge x limits=0.1,
	width=0.43\textwidth, 
	axis y line*=left,
	axis x line=none,
    ylabel style={yshift=-11pt},
	ylabel=Cache hit ratio,
    yticklabel style={font=\small,},
	ymin=0, ymax=1,
	ytick={0,0.2,0.4,0.6,0.8,1},
    symbolic x coords={0, 4, 8, 16, 32, 64, 128, 256, 512, 1024},
	xtick=data]

	\addplot[mark=oplus, mark options ={fill=red, draw=red}, samples=10, draw = red] coordinates {
            (4, 1)
		(8, 0.61)
		(16, 0.31)
		(32, 0.15)
		(64, 0.08)
		(128, 0.04)
		(256, 0.02)
		(512, 0.01)
            (1024, 0.005)
	};



    \end{axis}
	\end{tikzpicture}
	\vspace{-3mm}
    \caption{Cache hit ratio and amount of storage I/Os of Ginex~\cite{Par22} with varying storage I/O unit sizes.}
\label{fig:blockdatautil}
\vspace{-5mm}
\end{figure}
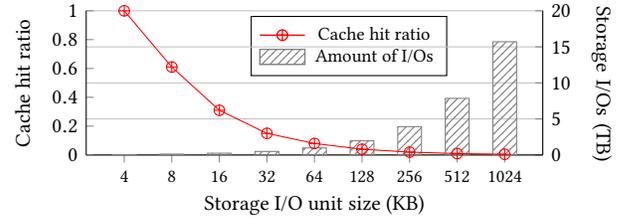

\subsection{Hyperbatch-based Processing}\label{subsec:proposed-performance}
\noindent
\textbf{Motivation.}
As mentioned in Section~\ref{sec:introduction}, handling a large number of small I/Os is a key challenge of storage-based GNN training.
We may tackle this challenge by increasing the size of a storage I/O unit and processing multiple units together, thereby improving the resource utilization.
However, the simple increase may lead to low efficiency of each storage I/O since a large amount of unnecessary data can be included in each I/O, which can waste the main memory space.
In order to evaluate the effect of increasing the size of a storage I/O unit,
we measure the amount of storage I/Os and relative data reuse of Ginex~\cite{Par22} with varying unit sizes on the PA dataset. 
Figure~\ref{fig:blockdatautil} shows that, as the size of storage I/O unit increases,
the total amount of storage I/Os (bar) grows, surpassing even 15 terabytes (TB), and the cache hit ratio decreases to below 0.06\%.
These results imply that simply increasing the storage I/O unit size is not a solution to this challenge.
In addition, due to the limited size of the main memory, 
the loaded data can be replaced with other data even though they are necessary for GNN training later.
In this case, the replaced data must be reloaded \textit{multiple times} from storage, which can incur an unnecessarily large amount of storage I/Os.



\vspace{1mm}
\noindent
\textbf{Key idea.}
From this motivation, 
we present a simple yet effective strategy to improve the efficiency of each storage I/O: 
\textit{hyperbatch-based} processing.
The idea behind this method is as follows:
if we process a single target node (minibatch) independently, it requires additional storage I/Os since many other nodes in the loaded block are not needed for the current target node.
Instead, processing nodes within the same block and reusing them across different target nodes (minibatches) together requires only a single block-wise storage I/O for each block.
In other words, our hyperbatch-based processing extends its processing scope from the target nodes in a single minibatch to those in multiple minibatches (we call it `hyperbatch' hereafter).


\begin{figure}[t]
		\centering
		\includegraphics[width=0.465\textwidth]{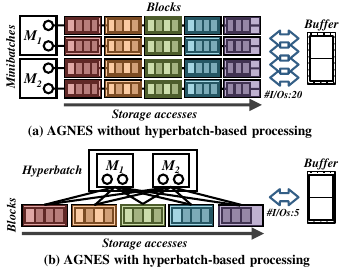}
		\vspace{-3mm}
		\caption{Effect of hyperbatch-based processing. It significantly reduces storage I/Os (e.g., 20 I/Os $\rightarrow$ 5 I/Os).}
	\vspace{-5mm}
	\label{fig:proposedtech}
\end{figure}

\vspace{1mm}
\noindent
\textbf{Example 1 (w/o hyperbatch-based processing).}
Figure~\ref{fig:proposedtech}(a)-(b) show an example of the data preparation process for four target nodes in two minibatches and five blocks with a buffer space of two blocks.
Each circle represents a target node, colored boxes represent the blocks containing the data (e.g., adjacency lists or feature vectors).
Here, blocks contain multiple data required by all target nodes.
Figure~\ref{fig:proposedtech}(a) illustrates the case when storage I/Os are performed from the perspective of target nodes (minibatches).
Whenever each block (e.g., a red box) containing neighboring nodes of the target node is processed,
the block may be replaced with other blocks (e.g., blue/purple boxes) required for processing other target nodes (minibatches).
In this case, the replaced block should be \textit{reloaded} from storage, resulting in a large number of storage I/Os (e.g., 20 storage I/Os).

\vspace{1mm}
\noindent
\textbf{Example 2 (w/ hyperbatch-based processing).}
Figure~\ref{fig:proposedtech}(b) illustrates the case when storage I/Os are performed from the perspective of blocks;
after a block (e.g., a red box) required for GNN training is loaded in main memory,
\textit{hyperbatch-based processing} processes neighboring nodes of multiple target nodes within the same minibatch and target nodes of a hyperbatch within a single iteration.
Consequently, this simple strategy significantly reduces additional storage I/Os (e.g., 5 storage I/Os). 

We will verify the effectiveness of hyperbatch-based processing on the performance of {\agnes} in Section~\ref{subsec:ablation-study}.

\subsection{Performance Consideration}\label{subsec:proposed-consideration}
In this section, we describe four key design considerations that affect the performance of {\agnes}.\footnote{All implementation details are available at {\codeurl}.}

\vspace{1mm}
\noindent
\textbf{(1) Sampling process.}
This process finds $k$-hop neighboring nodes of target nodes to be updated by a GNN model.
Each target node goes through multiple iterations corresponding to the number of hops (i.e., the number of GNN layers).
As the layer gets deeper, the number of neighbors grows significantly.
Since hyperbatch-based processing handles multiple target nodes in multiple minibatches at once,
it is likely that the same blocks are required in consecutive training iterations.
To further enhance the efficiency of each storage I/O, 
{\agnes} uses dynamic caching based on an LRU mechanism, also adopted in ~\cite{Han13,Jo19}, during the sampling stage, 
to pin graph blocks already in the graph buffer (e.g., the blocks processed in previous iterations) to prevent them from being replaced until they are completely processed in the current iteration.
{\agnes} unpins these blocks after they are completely processed.

\vspace{1mm}
\noindent
\textbf{(2) Gathering process.}
This process collects the sampled $k$-hop neighboring nodes of the target nodes and their feature vectors for transfer to the GPU.
The required feature vectors are moved to a \textit{contiguous} memory space in one iteration. 
Compared to graph topology, feature vectors require much larger storage space than graph topology~\cite{Par23,Sun23b}.
To efficiently use main memory space, 
{\agnes} counts the number of accesses to each feature vector and maintains only feature vectors whose access counts exceed a certain threshold, in a feature cache in main memory.
While the others (i.e., infrequently accessed feature vectors) are written back to storage at each minibatch and reloaded when they are required.

\vspace{1mm}
\noindent
\textbf{(3) Node identification.}
To manage the sampled nodes in different minibatches, 
we define a \textit{bucket}, $Bck$, which is a matrix containing the information of sampled nodes (i.e., their block IDs and minibatch IDs).
$Bck$ has rows and columns corresponding to the number of blocks and minibatches in a hyperbatch (i.e., the size of the hyperbatch), respectively.
Each cell of $Bck$, $Bck_{i,j}$, includes the nodes to be processed in the corresponding minibatch within a specific block.
Thus, {\agnes} identifies the nodes to be processed efficiently by scanning a row of the matrix, $Bck_{i,:}$.
Specifically, given target nodes to be processed, 
{\agnes} first (1) finds the index(es) of the block(s) containing the target nodes by referring to the object index table $T^g_{obj}$; (2) loads them into the main memory;
and (3) identifies the target nodes and their minibatch IDs by scanning $Bck_{i,:}$, the $i-{th}$ row of the bucket matrix that corresponds to the $i-{th}$ block.

\vspace{1mm}
\noindent
\textbf{(4) Asynchronous I/O.}
{\agnes} continuously loads parts of graph topology and feature vectors from storage to main memory, which are much slower than in-memory data transfers.
To achieve higher I/O-BW from \textit{more-frequent I/O requests} for the blocks to be processed,
{\agnes} adopts \textit{asynchronous} I/O processing.
After a thread issues an I/O request to the storage, 
the thread does not wait for the completion of the I/O in an idle state but rather tries to take over other tasks required to be processed.
This simple strategy could hide the costly I/O time within the overhead of other tasks,
thereby accelerating the process of data preparation.

\begin{algorithm}[t]
\DontPrintSemicolon 
\SetVlineSkip{0.0pt}
\SetInd{0.1em}{1.5em}
\SetKwInOut{Input}{\hspace{0.7em}Input}
\SetKwInOut{Output}{Output}
\Input{Topology $B_\textit{g}$, Feature $B_\textit{f}$, $T^\textit{g}_{\textit{obj}}$, Input nodes $N$}
\Output{Sampled nodes $N_{\textit{s}}$, Gathered feature vectors $F_{\textit{g}}$}
\SetKwFunction{FMain}{{\agnes}}
\SetKwFunction{FSub}{{\sf LoadData}}
\SetKwProg{Pn}{Function}{:}{\KwRet}
\Pn{\FMain{$B_\textit{g}$, $B_\textit{f}$, $T^\textit{g}_{\textit{obj}}$, input nodes $N$}}{
    $N_{\textit{s}}, F_{\textit{g}}, N_{\textit{in}}, T^\textit{g}_{\textit{buf}}, T^\textit{f}_{\textit{ch}} \leftarrow \emptyset$ \\
    $N^\textit{j}_{\textit{in}} \leftarrow N | N \in \textit{minibatch($\textit{j}$)}$ \\
    \For(\tcp*[f]{iterates by \# of layers}){$0, 1, ... , k$}{
        $N_{\textit{out}}, Bck \leftarrow \emptyset$ \\
        $Bck_{i,j} \leftarrow N^\textit{j}_{\textit{in}} | N^j_{\textit{in}} \in B_\textit{g}(\textit{i})$ \\ 
        \For(\tcp*[f]{sampling process}){$i, |Bck_{i,:}| \neq \emptyset$}{
            $\text{\sf LoadData}(\textit{i}, T^{\textit{g}}_{\textit{buf}}, T^{\textit{g}}_{\textit{obj}})$ \\
            \For(){$\textit{j}=0, 1, ... , |$\textit{minibatch}$|-1$} {
                $N^\textit{j}_{\textit{out}} \leftarrow N^\textit{j}_{\textit{out}} \cup \text{\sf Sample}(\textit{j}, Bck_{i,j}, T^\textit{g}_{\textit{buf}})$ \\
            }
        }
        $N_{\textit{s}} \leftarrow N_{\textit{s}} \cup N_{\textit{out}}$ \\
        $N_{\textit{in}} \leftarrow N_{\textit{out}}$ \\
    }
    $Bck \leftarrow \emptyset$, $Bck_{i,j} \leftarrow N_{\textit{s}} | N_{\textit{s}} \in B_\textit{f}(\textit{i})$ $\cap$ \textit{minibatch($\textit{j}$)} \\ 
    \For(\tcp*[f]{gathering process}){$\textit{i}, |Bck_{i,:}| \neq \emptyset$}{
        $\text{\sf LoadData}(\textit{i}, T^\textit{f}_{\textit{ch}})$ \\
        \For(){$\textit{j}=0, 1, ... , |$\textit{minibatch}$|-1$} {
            $F^\textit{j}_{\textit{g}} \leftarrow F^\textit{j}_{\textit{g}} \cup \text{\sf Gather}(\textit{j}, Bck_{i,j}, T^\textit{f}_{\textit{ch}})$ \\
        }
    }
    \Return{$N_{\textit{s}}, F_{\textit{g}}$}    \tcp*[f]{transfers to GPU}
}
\Pn{\FSub{$\textit{i}, T_{\textit{buf}}, T_{\textit{obj}}=\emptyset$}}{
    \For{$\textit{j}=0, 1, ... , |T_{\textit{obj}}|-1$}{
        $(n_1, n_2) \leftarrow T_{\textit{obj}}[\textit{j}]$ \\
        \If{$n_1 \leq \textit{i} \leq n_2$}{
            $\text{\sf b\_index} \leftarrow j$ \\
            \textbf{break}
        }
    }
    \If(\tcp*[f]{block-wise storage I/O}){$\text{\sf b\_index} \not\in T_{\textit{buf}}$}{
        $T_{\textit{buf}} \leftarrow T_{\textit{buf}} \cup \{\text{\sf b\_index}\}$ \\
    }
    \Return{}    
}
\caption{{\agnes}: {\sc Storage-based GNN}}\label{algo:agnes}
\end{algorithm}

\subsection{Algorithm of {\agnes}}\label{subsec:proposed-algo}
Algorithm~\ref{algo:agnes} shows the entire process of {\agnes} for performing GNN training.
Given a graph topology and node features, which are split and stored in multiple blocks with an object index table,
{\agnes} first loads the required adjacency lists and stores the sampled nodes in the sampling process (lines 3-12).
This process repeats the following steps in each iteration.
(1) the operation layer groups the nodes to be processed in the current iteration and stores the information about them in the corresponding cell of the bucket
by scanning the neighbor nodes sampled in the previous iteration for each minibatch;
(2) the operation layer requests the required graph blocks from the in-memory layer;
(3) the in-memory layer sends the block to the operation layer if it is already in the graph buffer, or otherwise, requests the block from the storage layer;
(4) then, the operation layer assigns the loaded block to a thread that processes the nodes by referring to the information stored in the bucket (i.e., hyperbatch-based processing) and stores the sampled nodes, which will be processed for the next iteration.
The gathering process is similar to the sampling process, except that {\agnes} loads the feature blocks and transfers the sampled nodes and their feature vectors to GPU (lines 13-18).

\input{figures/EQ1_single}
\section{Experimental Validation}\label{sec:experiment}
In this section, we comprehensively evaluate {\agnes} by answering the following evaluation questions (EQs):

\begin{itemize}	[leftmargin=10pt]
    \item \textbf{EQ1} (\textit{Training performance}). To what extent does {\agnes} improve existing GNN training methods?
    \item \textbf{EQ2} (\textit{Ablation study}). How does the hyperbatch-based processing contribute to improving the training performance of {\agnes}?
    \item \textbf{EQ3} (\textit{Sensitivity}). How sensitive is the performance of {\agnes} to different hyperparameter settings?
    \item \textbf{EQ4} (\textit{Accuracy}). How accurate is the GNN model trained by {\agnes}, compared to those trained by existing methods?
\end{itemize}

\subsection{Experimental Setup}
\noindent
\textbf{GNN models and datasets.}
We use three different GNN models: a 3-layer GCN~\cite{Kip16}, a 3-layer SAGE~\cite{Ham17}, and a 3-layer GAT~\cite{Vel18}.
We use their default parameters, following previous studies~\cite{Par22,Jia24} such as sampling size per layer as (10, 10, 10).
We set a block size as 1MB, a minibatch size as 1000, and a hyperbatch size as 1024.
To evaluate {\agnes}, we use five widely used real-world graph datasets -- IGB-medium (IG)~\cite{Kha23}, twitter-2010 (TW)~\cite{Jur14}, ogbn-papers100M (PA)~\cite{Hu20}, com-friendster (FR)~\cite{Jur14}, and yahoo-web (YH)~\cite{Yah24}. 
Table~\ref{table:datasets} shows the statistics of the datasets used in this experiment. 

\begin{table}[h]
\caption{Statistics of real-world graphs}
\vspace{-3mm}
\small
\centering
\setlength\tabcolsep{6pt} 
\begin{tabular}{c|c|c|c|c}
\toprule
Datasets & \#nodes & \#edges & Size (|F|=128) & Size (|F|=256) \\
\midrule
IG & 10M & 120M & 6GB & 11GB\\
TW & 41.65M & 1.47B & 32GB & 52GB \\ 
PA & 111.06M & 1.62B & 67GB & 120GB  \\ 
FR & 68.35M & 2.29B & 54GB & 87GB \\  
YH & 1.4B & 6.6B & 735GB & 1.4TB\\ 
\bottomrule
\end{tabular}
\label{table:datasets}
\end{table}

\vspace{1mm}
\noindent
\textbf{Competing methods.}
We compare {\agnes} with four state-of-the-art storage-based methods (Ginex~\cite{Par22}, GNNDrive~\cite{Jia24}, MariusGNN~\cite{Wal23}, and OUTRE~\cite{She24}) and one distributed GNN training method (DistDGL~\cite{Zhe20}).
We set the hyperparameters of existing methods as reported in their original works.
For Ginex, we set the superbatch size as 1024.

\vspace{1mm}
\noindent
\textbf{Evaluation protocol.}
We evaluate {\agnes} in terms of both efficiency (EQs 1-3) and accuracy (EQ 4).
For evaluation, we (1) train each of the three GNN models using each method on each dataset five times and (2) measure the average GNN training time per epoch (from sampling to training), along with the node classification accuracy.
To comprehensively evaluate {\agnes}'s capability to handle storage I/Os for graph topology and its feature vectors under a limited main memory environment, we consider two different memory settings: 
(1) Setting 1 (32\,GB): 16\,GB for the graph topology and 16\,GB for the feature vectors.
(2) Setting 2 (8\,GB): 4\,GB for the graph topology and 4\,GB for the feature vectors.

\vspace{1mm}
\noindent
\textbf{System configuration.}
We use a Dell R750 server equipped with the NVIDIA A40 GPU with 48\,GB device memory, two Intel Xeon Silver 4309Y CPU (each of which has 8 physical cores) with 128\,GB main memory, 
and PCIe Gen 4.0 NVMe SSDs as storage (each of which has a maximum I/O-BW of about 6.7\,GB/s).
We set the number of CPU threads as 16. 
We conduct all the experiments on Ubuntu 20.04.6, CUDA 12.1, Python 3.8.10, and PyTorch 2.2.

\subsection{EQ1. Training Performance}\label{subsec:performance-evaluation}
We compare the training performance of {\agnes} with those of four state-of-the-art storage-based and one distributed GNN training methods: Ginex~\cite{Par22}, MariusGNN~\cite{Wal23}, GNNDrive~\cite{Jia24}, OUTRE~\cite{She24}, and DistDGL~\cite{Zhe20}.
We consider two different buffer sizes for graph topology and node features:
(1) Setting 1 (32\,GB): this setting reflects a more-practical configuration commonly adopted in existing storage-based GNN training methods~\cite{Par22,Wal23,Jia24,She24}.
(2) Setting 2 (8\,GB): this setting is considered to rigorously evaluate the I/O handling capability of each method under constrained memory conditions, simulating the case where the graph size is significantly larger than main memory (i.e., I/O intensive setting).

\vspace{1mm}
\noindent
\textbf{Comparison with storage-based training methods.}
As shown in Figure~\ref{fig:EQ1_single}(a)–(b), {\agnes} consistently outperforms \textit{all} state-of-the-art storage-based GNN training methods across \textit{all} datasets.
Specifically, in Setting~1, {\agnes} achieves up to a 3.1$\times$ speedup over the best-performing competitor, Ginex.
In Setting~2, {\agnes} further widens the performance gap, outperforming Ginex up to 4.1$\times$.
This result indicates that {\agnes} can effectively handle storage I/O operations even under constrained memory conditions.
The superior performance of {\agnes} arises from its ability to mitigate a critical bottleneck in existing approaches (i.e., handling a large amount of small storage I/Os).
Existing methods issue a large number of small storage I/O requests upon cache misses, which prevents them from fully exploiting the I/O-BW provided by NVMe SSDs.
In contrast, although {\agnes} loads a larger amount of data via block-wise storage I/Os, it completes these I/O operations efficiently by fully utilizing the available NVMe I/O-BW.
As a result, the overhead of storage I/O is substantially reduced, leading to a significant improvement in training performance.

Note that N.A indicates a not-available case (e.g., MariusGNN and OUTRE support only the GraphSAGE model),
O.O.M denotes an out-of-memory case, as also reported in prior studies~\cite{Sun23b, Jia24},
and O.O.T represents an out-of-time case, where the entire execution, including preprocessing and training, requires more than 48 hours.
%


\begin{figure}[t]
	\centering
    \begin{tikzpicture}
	\begin{axis}[
	footnotesize,
	height=3.2cm,
	width=0.47\textwidth,
	axis x line*=center,
	axis y line*=left,
	ybar,
    bar width=10pt, 
	ymajorgrids=true,
	enlarge x limits=0.1, 
    major grid style={line width=.2pt,draw=gray!50},
	xlabel style={yshift=0pt},
    ylabel style={yshift=-9pt, align=center,},
    ylabel=Exec. time (sec.),
    ymin=0, ymax=250,
    ytick={0,50, 100, 150, 200, 250},
    scaled y ticks = false,
    yticklabel style={font=\footnotesize,},
	symbolic x coords={1, 2, 4, 8, 16, },
    xtick=data,
    xticklabel style={font=\footnotesize,},
	every node near coord/.append style={color=black, font=\footnotesize},
    after end axis/.code={
        \draw [dashed, line width=0.8pt, black, decoration={, amplitude=1pt}, decorate] (rel axis cs:0.827,-0.35) -- (rel axis cs:0.827,1);
    },
    ]
    \addplot+[ybar, pattern = north east lines, area legend, pattern color = black, draw=black, 	point meta=y,	visualization depends on=rawy\as\rawy, nodes near coords={\pgfmathprintnumber{\rawy}}] 
	coordinates {
        (1, 200)
        (2, 101)
        (4, 49)
        (8, 25)
        (16, 13)
        (, 113)
    };
\end{axis}
\node[] at (3.1, -0.55) {\small DistDGL~\cite{Zhe20}};
\node[] at (6.218, -0.3) {\footnotesize 1};
\node[] at (6.218, -0.54) {\small {\agnes}};
\node[] at (-0.6, -0.28) {\footnotesize \#machines};
\end{tikzpicture}
\vspace{-3mm}
    \caption{Comparison with a distributed training method: {\agnes} achieves comparable training performance to a distributed method only with limited computing resources.}
\label{fig:EQ1_dist}
\vspace{-5mm}
\end{figure}
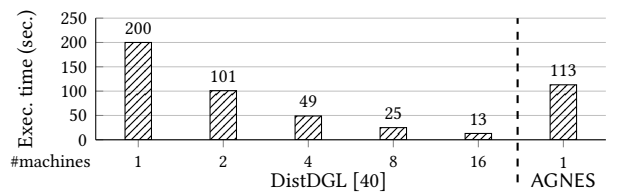

\vspace{1mm}
\noindent
\textbf{Comparison with a distributed training method.}
In addition, we evaluate {\agnes} against a distributed GNN training approach, DistDGL~\cite{Zhe20}. DistDGL was evaluated on a cluster of 16 AWS m5.24xlarge instances, each equipped with 96 vCPUs, 384\,GB memory, and interconnected via a 100\,Gbps network~\cite{Zhe20}.
Since replicating such a high-end distributed environment is infeasible, we quote the performance results on the PA dataset as reported in~\cite{Zhe20}.

Figure~\ref{fig:EQ1_dist} shows that {\agnes} achieves performance comparable to DistDGL running on two instances, despite being executed on a \textit{single machine with limited computational resources}.
Considering that DistDGL eliminates storage I/Os by maintaining the \textit{entire graph in large main memory} across high-end distributed nodes, 
this result highlights the efficiency of {\agnes} in enabling large-scale GNN training with substantially lower infrastructure requirements.
This advantage arises since {\agnes} incurs only intra-machine communication overhead (i.e., storage I/Os), 
which is significantly smaller than the inter-machine communication overhead inherent to distributed systems.
Moreover, this comparison demonstrates that carefully optimized storage-based GNN training can narrow the performance gap with distributed training approaches.
As a result, {\agnes} provides a \textit{practical} and \textit{cost-effective} alternative for large-scale GNN training, especially in scenarios where access to expensive distributed clusters is limited or impractical.
These findings further validate the scalability of {\agnes} for large-scale GNN training under realistic resource constraints.

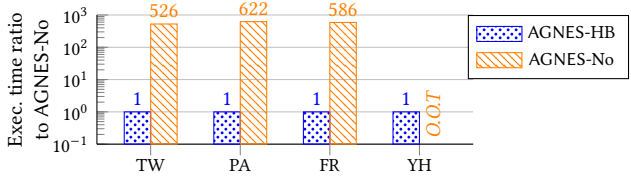
\begin{figure}[]
\begin{tikzpicture}
    \begin{axis}[
	height=3.3cm,
	width=0.37\textwidth,
	ymajorgrids=true,
	major grid style={line width=.2pt,draw=gray!50},
	axis x line*=center,
	axis y line*=left,
	ybar=0pt,
    ymode=log,
    log origin=infty,
	bar width=10pt,
	enlarge x limits=0.2,
	title style={yshift=-90pt,font=\normalsize},
	xlabel style={yshift=0pt,font=\scriptsize},
	ylabel style={align=center, yshift=-9pt,font=\small},
	ylabel= Exec. time ratio\\to {\agnes}-No,
	ymin=0.1,
	ymax=1000,
	ytick={0.1, 1, 10, 100, 1000},
	symbolic x coords={TW,PA,FR,YH},
	xtick=data,
	xticklabel style={draw=none,font=\footnotesize,yshift=2pt},
	yticklabel style={font=\footnotesize},
	legend style={at={(1.2,1)},anchor=north, draw=black},
	legend cell align=center,
	legend columns=1,
	legend style={font=\footnotesize},
    every node near coord/.append style={font=\small, anchor=center,yshift=5pt},
	]
    \addplot[font=\footnotesize, blue, pattern = crosshatch dots, area legend, pattern color = blue, 	point meta=y,	visualization depends on=rawy\as\rawy, nodes near coords={\pgfmathprintnumber{\rawy}}]
    coordinates {
		(TW, 1)
		(PA, 1)
        (FR, 1)
        (YH, 1)
	};\addlegendentry{{\agnes}-HB}
    \addplot[font=\footnotesize, orange, pattern = north west lines, area legend, pattern color = orange, point meta=y,	visualization depends on=rawy\as\rawy, nodes near coords={\pgfmathprintnumber{\rawy}}]
    coordinates {
        (TW,526)
        (PA,622)
        (FR,586)
	};\addlegendentry{{\agnes}-No}

	\end{axis}
	\node[font=\small,color=orange, rotate=90] at (4.47,0.4) {\textit{O.O.T}};
    \end{tikzpicture}

    \footnotesize
    \centering
	\vspace{-4mm}
    \caption{Effect of the hyperbatch-based processing on the performance of {\agnes}. It significantly improves the training performance of {\agnes}.}
	\vspace{-3mm}
	\label{fig:EQ2_ablationoverall}
\end{figure}

\begin{figure}[]
	\centering
	\begin{tikzpicture}
	\begin{axis}[
	footnotesize,
	height=3.5cm,
	width=0.22\textwidth, 
	axis x line*=center,
	axis y line*=left,
	ymajorgrids=true,
	enlarge x limits=0.15, 
    major grid style={line width=.2pt,draw=gray!50},
	xlabel style={yshift=2pt},
    xlabel = (a) Block size (KB),
    ylabel style={yshift=-16pt, align=center,},
    ylabel=Exec. time (sec.),
    ymin=0, ymax=2000,
    ytick={0, 1000, 2000},
    yticklabels={0,1k,2k},
    scaled y ticks = false,
    yticklabel style={font=\small,},
	symbolic x coords={64, 128, 256, 512, 1024, 2048, 4096},
    xtick=data,
    xticklabel style={font=\footnotesize,rotate=45,yshift=3pt,xshift=1pt},
    legend style={at={(0.5,1.6)}, anchor=north, draw=none, legend columns=-1},
    every node near coord/.append style={
        color=black, font=\scriptsize, yshift=-1.5pt
    }
	]	
    \addplot[mark=square, blue, line width=0.7pt,	mark options={mark size=2pt}, samples=10] coordinates {
		(64, 1710)
		(128, 1571)
		(256, 1441)
		(512, 1288)
        (1024, 1228)
        (2048, 1315)
        (4096, 1524)
	};

	\end{axis}

	\begin{axis}[
	footnotesize,
	height=3.5cm,
	enlarge x limits=0.15,
	width=0.22\textwidth, 
	axis y line*=right,
	axis x line=none,
    ylabel style={anchor=center,rotate=180,yshift=120pt},
	ylabel= \# of storage I/Os (M),
	ymin=0, ymax=80,
	ytick={0,20,40,60,80},
    symbolic x coords={64, 128, 256, 512, 1024, 2048, 4096},
	xtick=data]

    \addplot[mark=o, red, line width=0.7pt,	mark options={mark size=2pt}, samples=10] coordinates {
		(64, 63)
		(128, 34)
		(256, 19)
		(512, 10)
        (1024, 5)
        (2048, 3)
        (4096, 2)
	};


    \end{axis}
	\end{tikzpicture}
	\begin{tikzpicture}
	\begin{axis}[
	footnotesize,
	height=3.5cm,
	width=0.22\textwidth, 
	axis x line*=center,
	axis y line*=left,
	ymajorgrids=true,
	enlarge x limits=0.15, 
    major grid style={line width=.2pt,draw=gray!50},
	xlabel style={yshift=2pt},
    xlabel = (b) Hyperbatch size,
    ylabel style={yshift=-16pt,align=center,},
    ylabel=Exec. time (sec.),
    ymin=0, ymax=8000,
    ytick={0, 2000, 4000, 6000, 8000},
    yticklabels={0,2k,4k,6k,8k},
    scaled y ticks = false,
    yticklabel style={font=\small,},
	symbolic x coords={64, 128, 256, 512, 1024, 2048},
    xtick=data,
    xticklabel style={font=\footnotesize,rotate=45,yshift=3pt,xshift=1pt},
    legend style={at={(0.5,1.6)}, anchor=north, draw=none, legend columns=-1},
    every node near coord/.append style={
        color=black, font=\scriptsize, yshift=-1.5pt
    }
	]	
    \addplot[mark=square, blue, line width=0.7pt,	mark options={mark size=2pt}, samples=10] coordinates {
		(64, 6400)
		(128, 5100)
		(256, 3360)
		(512, 1963)
        (1024, 1228)
        (2048, 1291)
	};

	\end{axis}
	\begin{axis}[
	footnotesize,
	height=3.5cm,
	enlarge x limits=0.15,
	width=0.22\textwidth, 
	axis y line*=right,
	axis x line=none,
    ylabel style={anchor=center,rotate=180,yshift=120pt},
	ylabel= \# of storage I/Os (M),
	ymin=0, ymax=80,
	ytick={0,20,40,60,80},
    symbolic x coords={64, 128, 256, 512, 1024, 2048, 4096},
	xtick=data]

    \addplot[mark=o, red, line width=0.7pt,	mark options={mark size=2pt}, samples=10] coordinates {
		(64, 48)
		(128, 31)
		(256, 18)
		(512, 10)
        (1024, 6)
        (2048, 3)
	};


    \end{axis}
	\end{tikzpicture}
	\vspace{-4mm}
	\caption{Execution time (blue) and the number of storage I/Os (red) according to block size and hyperbatch size.}
\label{fig:EQ2_ablationparameter}
\vspace{-3mm}
\end{figure}
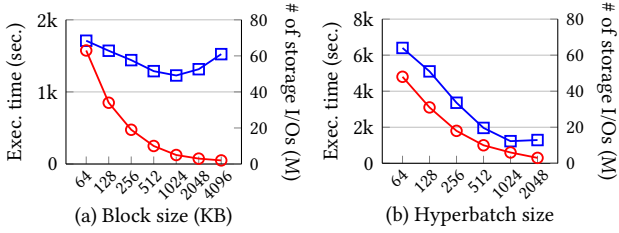

\subsection{EQ2. Ablation Study}\label{subsec:ablation-study}
Then, we evaluate the effectiveness of the hyperbatch-based processing.
We compare the two following versions of {\agnes}:
\begin{itemize}	[leftmargin=10pt]
    \item {\agnes}-No: {\agnes} without hyperbatch-based processing.
    \item {\agnes}-HB: {\agnes} with hyperbatch-based processing.
\end{itemize}

Figure~\ref{fig:EQ2_ablationoverall} shows the ratio of the execution time of {\agnes}-No compared to that of {\agnes}-HB.
The results show that our proposed strategy significantly enhances GNN training performance.
Specifically, the hyperbatch-based processing improves the execution time of {\agnes} by up to $622\times$ by significantly increasing the efficiency of each storage I/O.
This improvement arises because the hyperbatch-based processing effectively eliminates a large number of redundant and small storage I/Os by aggregating them into fewer, larger, and more contiguous I/O requests.
As a result, the overhead caused by frequent small storage I/Os can be greatly reduced, allowing {\agnes} to fully utilize the I/O-BW.
These results suggest that mitigating excessive small storage I/Os is as critical as efficiently exploiting in-memory caches, especially in storage-based GNN training methods.
Note that O.O.T. denotes an out-of-time case, where execution requires more than 24 hours.

We also conduct experiments to validate the effects of block size and hyperbatch size using the YH dataset, the largest dataset in our experiments.
We measure (1) the number of storage I/Os and (2) the total execution time by varying the block size from 64KB to 4096KB and the hyperbatch size from 64 to 2048.
Figure~\ref{fig:EQ2_ablationparameter}(a)–(b) shows the results, where the $x$-axis represents the block/hyperbatch size, and the $y$-axis represents the total execution time (left) and the total number of storage I/Os (right).
First, {\agnes} achieves the best performance when the block size is 1024KB.
Thus, although the number of storage I/Os decreases as the block size increases, the proportion of unnecessary data fetched within each block also increases.
Second, {\agnes} shows the best performance when the hyperbatch size is larger than 1024.
This result indicates that, as the hyperbatch size increases, the number of storage I/Os required across multiple minibatches decreases, while the overhead of hyperbatch-based processing gradually increases.





\input{figures/EQ3}

\subsection{EQ3. Sensitivity Analysis}\label{subsec:sensitivity-analysis}
We evaluate how sensitive the performance of {\agnes} is to hyperparameters: (1) buffer size, (2) the number of CPU threads, (3) feature dimension, (4) sampling size, and (5) SSD array size.

\vspace{1mm}
\noindent
\textbf{(1) Buffer size.}
Figure~\ref{fig:EQ3}(a) shows the performance of {\agnes} and Ginex with varying buffer sizes from 1\,GB to 16\,GB,
where the $x$-axis denotes the buffer size and the $y$-axis denotes the execution time.
The execution time of Ginex increases rapidly as the buffer size decreases.
This result indicates that a smaller buffer size leads to a large number of small storage I/Os, thereby significantly degrading overall performance.
In contrast, the performance of {\agnes} remains stable across different buffer sizes,
indicating that {\agnes} efficiently utilizes data within each block and thus substantially reduces the number of storage I/Os.
Notably, even when the buffer size increases to 16\,GB for the TW and PA datasets, the execution time of Ginex also increases.
This is because Ginex requires a considerable amount of time to load the entire cache into main memory before starting the data preparation process.
Therefore, these results demonstrate that the performance of {\agnes} is less sensitive to variations in main memory size.

\vspace{1mm}
\noindent
\textbf{(2) Number of CPU threads.}
Figure~\ref{fig:EQ3}(b) shows the performance of {\agnes} and Ginex with varying numbers of CPU threads from 1 to 16,
where the $x$-axis denotes the number of CPU threads and the $y$-axis denotes the execution time.
The execution time of both methods decreases as the number of threads increases.
Notably, {\agnes} achieves a larger performance improvement than Ginex as the number of threads increases.
This result indicates that {\agnes} utilizes multiple CPU threads more effectively during the data preparation stage,
thereby better exploiting parallelism compared to Ginex.


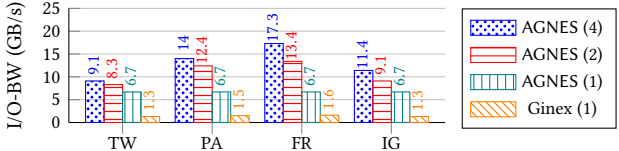
\begin{figure}[]
\begin{tikzpicture}
    \begin{axis}[
	height=3.1cm,
	width=0.37\textwidth,
	ymajorgrids=true,
	major grid style={line width=.2pt,draw=gray!50},
	axis x line*=center,
	axis y line*=left,
	ybar=0pt,
	bar width=7pt,
	enlarge x limits=0.2,
	title style={yshift=-90pt,font=\normalsize},
	xlabel style={yshift=0pt,font=\scriptsize},
	ylabel style={align=center, yshift=-15pt,font=\small},
	ylabel=I/O-BW (GB/s),
	ymin=0,
	ymax=25,
	ytick={0, 5, 10, 15, 20, 25},
	symbolic x coords={TW,PA,FR,IG},
	xtick=data,
	xticklabel style={draw=none,font=\footnotesize,yshift=2pt},
	yticklabel style={font=\footnotesize},
	legend style={at={(1.25,1)},anchor=north, draw=black},
	legend cell align=center,
	legend columns=1,
	legend style={font=\footnotesize},
    every node near coord/.append style={rotate=90,font=\scriptsize, anchor=center,xshift=6pt,yshift=0.5pt},
	]
    \addplot[font=\footnotesize, blue, pattern = crosshatch dots, area legend, pattern color = blue, 	point meta=y,	visualization depends on=rawy\as\rawy, nodes near coords={\pgfmathprintnumber{\rawy}}]
    coordinates {
        (TW,9.1)
        (PA,14.0)
        (FR,17.3)
        (IG,11.4)
	};\addlegendentry{{\agnes} (4)}
    \addplot[font=\footnotesize, red, pattern = horizontal lines, area legend, pattern color = red, point meta=y,	visualization depends on=rawy\as\rawy, nodes near coords={\pgfmathprintnumber{\rawy}}]
    coordinates {
        (TW,8.3)
        (PA,12.4)
        (FR,13.4)
		(IG,9.1)
	};\addlegendentry{{\agnes} (2)}
	\addplot[font=\footnotesize, teal, pattern = vertical lines, area legend, pattern color = teal, point meta=y,	visualization depends on=rawy\as\rawy, nodes near coords={\pgfmathprintnumber{\rawy}}]
	coordinates {
        (TW,6.7)
        (PA,6.7)
        (FR,6.7)
        (IG,6.7)
	};\addlegendentry{{\agnes} (1)}
    \addplot[font=\footnotesize, orange, pattern = north west lines, area legend, pattern color = orange, point meta=y,	visualization depends on=rawy\as\rawy, nodes near coords={\pgfmathprintnumber{\rawy}}]
    coordinates {
		(TW,1.3)
		(PA,1.5)
        (FR,1.6)
        (IG,1.3)
	};\addlegendentry{Ginex (1)}
	\end{axis}
    \end{tikzpicture}

    \footnotesize
    \centering
	\vspace{-4mm}
    \caption{The maximum I/O-BW utilization of {\agnes} and Ginex according to the number of SSDs.}
	\vspace{-4mm}
	\label{fig:EQ3_sub}
\end{figure}

\vspace{1mm}
\noindent
\textbf{(3) Feature dimension.}
Figure~\ref{fig:EQ3}(c) shows the performance of {\agnes} and Ginex with varying feature dimensions from 64 to 512,
where the $x$-axis denotes the feature dimension and the $y$-axis does the execution time.
The execution time of {\agnes} increases as the feature dimension increases.
Notably, while {\agnes} is consistently faster than Ginex, the performance improvement is more pronounced at smaller feature dimensions.
This is because, when the feature dimension is smaller, {\agnes} can retrieve more feature vectors with a single block-wise storage I/O.
In contrast, Ginex's small storage I/O, with a minimum size of 4\,KB, results in poor data utilization for smaller feature vectors.

\vspace{1mm}
\noindent
\textbf{(4) Sampling size.}
Figure~\ref{fig:EQ3}(d) shows the performance of {\agnes} and Ginex with varying sampling sizes from 5 to 15,
where the $x$-axis denotes the sampling size per layer and the $y$-axis does the execution time.
The execution time of {\agnes} increases linearly as the sampling size increases.
As the sampling size increases, Ginex experiences a significant increase in small storage I/Os, since more $k$-hop neighboring nodes are required for GNN training.
In contrast, {\agnes} benefits from the reduced number of block-wise storage I/Os, thereby improving its performance.

\vspace{1mm}
\noindent
\textbf{(5) SSD array size.}
Figure~\ref{fig:EQ3}(e) shows the performance of {\agnes} and Ginex with varying SSD array sizes from 1 to 4\footnote{We configure RAID0 with Linux mdadm to utilize I/O-BW provided by all NVMe SSDs in parallel.},
where the $x$-axis denotes the SSD array size and the $y$-axis does the execution time.
{\agnes} reduces the overall execution time by approximately 18\% on average and achieves up to a 27\% reduction on the IG dataset.
On the other hand, the execution time of Ginex remains unchanged even as the SSD array size increases.
This is because a large number of small storage I/Os prevents Ginex from fully utilizing the I/O-BW of even a single NVMe SSD.
Figure~\ref{fig:EQ3_sub} shows the maximum I/O-BW utilization of {\agnes} and Ginex, where the $x$-axis denotes datasets and the $y$-axis does I/O-BW utilization.
The results show that {\agnes} is able to fully utilize the I/O-BW provided by multiple NVMe SSDs by up to 17.3\,GB/s.

\input{figures/EQ4_acc}

\subsection{EQ4. Accuracy}\label{subsec:accuracy}
Finally, we evaluate the performance of {\agnes} in terms of accuracy per training time.
We train three GNN models on the PA and IG datasets using {\agnes} and Ginex for 10 epochs, and measure their node classification accuracy at every epoch.
Figure~\ref{fig:EQ4_accplus} shows the results, where the $x$-axis denotes the elapsed time and the $y$-axis does the accuracy.
The results show that {\agnes} achieves the same accuracy as Ginex at every epoch, regardless of the dataset and GNN model.
However, {\agnes} reaches the same accuracy in a shorter amount of time, resulting in a higher accuracy improvement per unit time compared to Ginex.
This indicates that {\agnes} enables faster convergence while preserving model accuracy.
Overall, these results demonstrate that {\agnes} is a more effective solution for large-scale GNN training,
as it substantially improves training efficiency without sacrificing accuracy.

\vspace{-1.5mm}
\section{Conclusions}\label{sec:conclusion}
In this paper,
we observe that the existing storage-based GNN training methods suffer from a serious bottleneck of data preparation since they have overlooked a natural yet critical challenge: 
\textit{how to handle a large number of small storage I/Os}.
To address this challenge, we propose a novel storage-based approach to GNN training (\textbf{{\agnes}}) with the 3-layer architecture to handle storage I/Os efficiently.
We identify important issues causing serious performance degradation in storage-based training and propose a simple yet effective strategy, hyperbatch-based processing, that improves the efficiency of storage I/Os based on the unique characteristics of real-world graphs.
Through extensive experiments using five web-scale graphs, 
we demonstrate that {\agnes} significantly outperforms state-of-the-art storage-based GNN training methods in terms of accelerating large-scale GNN training. 


\vspace{-1.5mm}
\begin{acks}
This is a joint work between Samsung Electronics Co., Ltd, and Hanyang University (No. IO251222-14891-01);
the authors also would like to thank SMRC (Samsung Memory Research Center) for providing the infrastructure for this work.
This work was supported by the Institute of Information \& Communications Technology Planning \& Evaluation (IITP) grant funded by the Korea government (MSIT) (No. RS-2020-II201373 and No. RS-2022-00155586).
The work of Yunyong Ko was supported by the National Research Foundation of Korea (NRF) grant, funded by the Korea government (MSIT) (No. RS-2024-00459301).
\end{acks}

\clearpage

\bibliographystyle{ACM-Reference-Format}
\balance
\bibliography{references}


\end{document}